\documentclass{ifacconf}

\makeatletter
\let\old@ssect\@ssect % Store how ifacconf defines \@ssect
\makeatother

\usepackage{natbib}
\usepackage{hyperref}

\makeatletter
\def\@ssect#1#2#3#4#5#6{%
  \NR@gettitle{#6}% Insert key \nameref title grab
  \old@ssect{#1}{#2}{#3}{#4}{#5}{#6}% Restore ifacconf's \@ssect
}
\makeatother

\usepackage{graphicx}      % include this line if your document contains figures
\usepackage{amsmath}
\usepackage{amssymb}
\usepackage{algorithmicx}
\usepackage{algorithm}
\usepackage{algcompatible}

\usepackage{multirow}

\begin{document}
\begin{frontmatter}

\title{SA-reCBS: Multi-robot task assignment with integrated reactive path generation\thanksref{footnoteinfo}} 

\thanks[footnoteinfo]{This work has been partially funded by the European Unions Horizon 2020 Research and Innovation Programme under the Grant Agreements No. 101003591 NEX-GEN SIMS}

\author{Yifan Bai},
% \author[label2]{Yu Wang}
\author{Christoforos Kanellakis} and 
\author{George Nikolakopoulos}

\address{Robotics and AI Team\\ Lule{\aa} University of Technology, Sweden}
% \address[label2]{Applied Mathematics\\
%  Xi'an Jiaotong-Liverpool University, China
%  }
% \address[Second]{Colorado State University, 
%    Fort Collins, CO 80523 USA (e-mail: author@lamar. colostate.edu)}
% \address[Third]{Electrical Engineering Department, 
%    Seoul National University, Seoul, Korea, (e-mail: author@snu.ac.kr)}

\begin{abstract}                % Abstract of not more than 250 words.
In this paper, we study the multi-robot task assignment and path-finding problem (MRTAPF), where a number of robots are required to visit all given tasks while avoiding collisions with each other. We propose a novel two-layer algorithm SA-reCBS that cascades the simulated annealing algorithm and conflict-based search to solve this problem. Compared to other approaches in the field of MRTAPF, the advantage of SA-reCBS is that without requiring a pre-bundle of tasks to groups with the same number of groups as the number of robots, it enables a part of robots needed to visit all tasks in collision-free paths. We test the algorithm in various simulation instances and compare it with state-of-the-art algorithms. The result shows that SA-reCBS has a better performance with a higher success rate, less computational time, and better objective values.

\end{abstract}


\end{frontmatter}
%===============================================================================

\section{Introduction}
\begin{figure}[!htbp]
\centering
    \includegraphics[width = .9\linewidth]{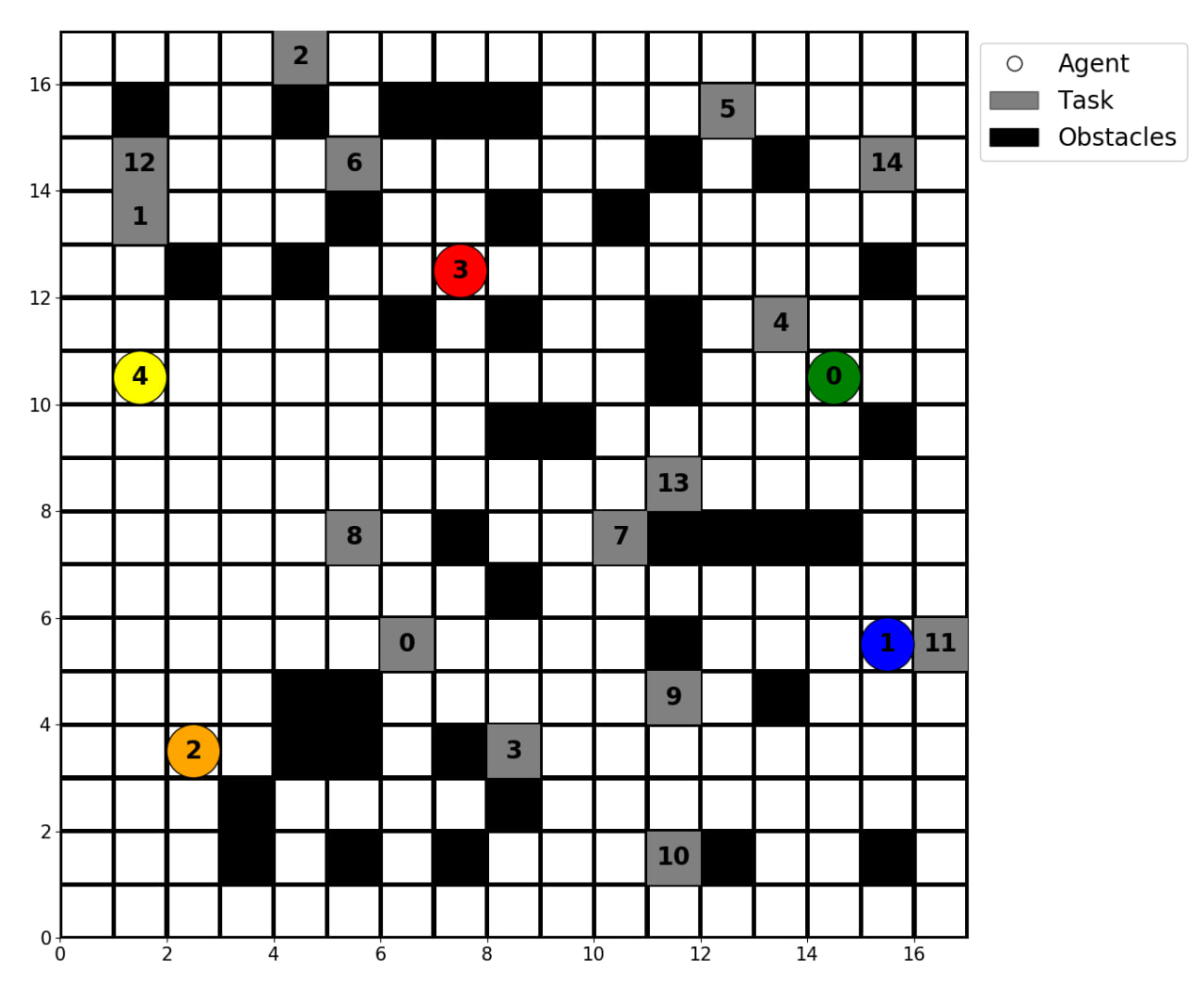}
\caption{Example of MRTAPF with 5 robots and 15 tasks, where circles, grey blocks, and black blocks represent start locations of robots, tasks, and obstacles respectively.}
\label{fig:MRTAPF}
\end{figure}

Taking advantage of conducting collective behaviors that may offer high-efficiency, redundancy, and robustness, multi-robot systems (MRS) have attracted attention from the scientific community. Nowadays, MRS have been applied in many pertinent areas of industry such as surveillance and monitoring (\citeauthor{li2014multi}, \citeyear{li2014multi}), search and rescue (\citeauthor{queralta2020collaborative}, \citeyear{queralta2020collaborative}), logistics (\citeauthor{farinelli2017advanced}, \citeyear{farinelli2017advanced}), etc.\\
The fundamental problem of MRS is to allocate tasks to robots and find conflict-free paths for the robots whilst optimizing an objective.
In this paper, we study a generic task assignment and path-finding problem for MRS in a known obstacle-ridden environment. There are different numbers of robots and tasks. The problem is to assign the tasks to the desired number of robots that start from different locations and plan collision-free paths for the robots such that all tasks are visited and the flowtime(the sum of all robots' travel time) is minimized. We call this problem multi-robot task assignment and path-finding (MRTAPF), while Figure~\ref{fig:MRTAPF} depicts an illustrative scenario.

\subsection{Related Work}
Just as its name implies, the MRTAPF problem consists of two components: Multi-robot task assignment (MRTA) and Multi-agent path-finding (MAPF), both of which have been extensively researched in the past. In recent years, some researchers have put effort into solving task assignment and path-finding problems simultaneously. In this section, we provide a brief review of existing approaches to tackle these problems.\\

\textbf{MRTA:} The MRTA problem addresses the issue of assigning tasks to robots such that an overall system goal is achieved. \cite{khamis2015multi} provided a review of approaches to solving the MRTA problems. One main approach is the auction-based algorithm, where robots use a negotiation protocol to bid for tasks. There could be a central auctioneer who decides which task to which agent with global knowledge, or in a distributed manner, robots have local information of the ambient environment and communicate with each other instead of the central coordinator. \cite{choi2009consensus} proposed the consensus-based bundle algorithm (CBBA), which allows robots to bid on a task asynchronously based on its own situational awareness and then applies consensus strategy to converge the winning bids of robots.\\
Apart from auction-based methods, optimization-based techniques are also widely used, composed of deterministic and stochastic techniques. Deterministic techniques consist of graph-based methods, mixed integer linear programming (MILP), the Hungarian algorithm, etc. Stochastic methods contain trajectory-based algorithms such as simulated annealing algorithm (\citeauthor{xiao2021simulated}, \citeyear{xiao2021simulated}), population-based algorithms like genetic algorithm (\citeauthor{zhu2018multi}, \citeyear{zhu2018multi}), ant colony optimization (\citeauthor{chen2022efficient}, \citeyear{chen2022efficient}) and some hybrid optimization approaches (\citeauthor{xue2021hybrid}, \citeyear{xue2021hybrid}).  \\
\\
\textbf{MAPF:} The MAPF problem aims to plan paths for multiple agents to navigate from their starts to destinations without colliding with each other. \citeauthor{yu2013structure} (\citeyear{yu2013structure}) proved that the MAPF problem is NP-hard. Prioritized planning (\citeauthor{prioritizedplanning}, \citeyear{prioritizedplanning}) plans paths for each agent in order of their priority. The planned path of an agent is required to avoid conflicts with paths already found for agents with higher priority. Some extensions of $A^*$ algorithm are able to solve MAPF optimally. For instance, $M^*$ algorithm, developed by \cite{wagner2015subdimensional}, is an implementation of subdimensional expansion over $A^*$, constructing a variable dimensional search space to resolve conflicts when necessary. \cite{grenouilleau2019multi} developed a Multi-label $A^*$ that adds labels indicating the order of a list of goals and computes the shortest path for an agent to visit them.
One of the state-of-the-art techniques, namely conflict-based search (CBS) (\citeauthor{sharon2015conflict}, \citeyear{sharon2015conflict}), performs a search on a conflict tree based on conflicts between individual robots at the high-level, and at the low-level plans path for a single agent that satisfy the constraints imposed by the high-level node. CBS is proved to be optimal and complete.\\
\\
\textbf{MRTAPF:} Multi-robot task-assignment and path-finding problems are the integration of MRTA and MAPF. The CBS-TA (\citeauthor{honig2018conflict}, \citeyear{honig2018conflict}) framework builds upon CBS to identify assignment solutions within a search forest. However, it has a limitation of being able to assign only one exact goal to each agent, even if there are additional unvisited goals remaining. \cite{zhong2022optimal} combines CBS-TA and MLA* to achieve a one-task-to-one-agent assignment, and each task consists of a sequence of goals rather than a single goal. However, the goals are grouped and sequenced originally. MS* (\citeauthor{ren2021ms}, \citeyear{ren2021ms}) is an exact algorithm that embeds mTSP solver to the M* algorithm to address goal sequencing and path-finding simultaneously. Multi-agent pick-up and delivery (MAPD), in the vein of TAPF, is a more complicated problem in which each task has two ordered goals (\citeauthor{brown2020optimal}, \citeyear{brown2020optimal}; \citeauthor{xu2022multi}, \citeyear{xu2022multi}).

\subsection{Contributions}
Based on the current SoA, this work addresses the low execution time and low success rate issues when solving MRTAPF with larger quantities of robots and tasks.
 We propose SA-reCBS, a two-stage algorithm for MRTAPF problems. The first stage assigns tasks to robots by solving a multi-depot vehicle routing problem by means of a simulated annealing algorithm, assuming that robots will not collide. The second stage is responsible for researching the collision-free path sets. It takes the assignment result from the first level and iteratively calls CBS algorithm during the course of approaching tasks in the predetermined order, until all the tasks are visited. In addition, we evaluate the computational time of the proposed scheme on several instances, with a combination of a number of robots and tasks on different maps. The results indicate that SA-reCBS outperforms the state-of-the-art multi-goal sequencing and path-finding algorithm (\citeauthor{ren2022conflict}, \citeyear{ren2022conflict}) in terms of success rate, computational time, and total cost. 
\section{Problem Description}
Consider an undirected graph $G=\left(V,E\right)$, where $v \in V$ corresponds to locations and $e \in E$ are unit-weight edges connecting locations. There are $N$ robots $\left\{a_1,a_2,\ldots,a_n\right\}$ starting at different locations $s_i \in V, i \in \left\{ 1,2,\ldots,n \right\}$, with a set of $M$ different goal locations $\left\{g_1,g_2,\ldots,g_m\right\} \subseteq V$ to be visited.\\
All robots share a global clock. At each time step, an robot can either wait at its current vertex or move along an edge to an adjacent vertex. We denote the location of robot $i$ at time $t$ as $\pi_i^t$, thus a path for robot $a_i$ is a sequence of locations $\Pi_i = \left( \pi_i^0,\ldots,\pi_i^t,\ldots\right)$. The aim of this paper is to find paths for all the robots such that:\\
(1) Each robot starts at its starting locations and terminates at one of the goal locations;\\
(2) Each goal location is visited exactly once by a robot;\\
(3) The planned paths of robots must be conflict-free;\\
% (4) The maximal makespan $\tau$, which is the time when the last goal is visited, is minimized.\\
(4) The total travel time of each robot is minimized.\\
Two robots are considered to be in conflict in two cases: (a) \textbf{vertex conflict}, where two robots occupy the same vertex at the same time, that is $u=\pi^t_i=\pi^t_j$, or\\ (b) \textbf{edge conflict}, where two robots traverse the same edge from opposite directions at time $t$, that is $u=\pi^t_i=\pi^{t+1}_j$ and $v=\pi^{t+1}_i=\pi^t_j$.\\
% Noticeable, as the path $\Pi_i$ of robot $a_i$ is a time-space sequence, the makespan $\tau$ is equivalent to the maximum length of paths among all robots.

\section{Methodology}
In this section, we present a two-stage algorithm called SA-reCBS to tackle the multi-robot task assignment and path-finding problem, where the conflict-based search is leveraged to plan conflict-free paths for each robot to move towards their assigned goal locations sequentially provided by the simulated annealing algorithm. 

\subsection{Simulated Annealing}
Ignoring the possible conflicts between robots, the assignment level takes the starting locations of the robots, the goal locations, and the map information as inputs and outputs visiting sequences of goal locations for all robots. This procedure can be formulated as a simplified Multi-Depot Vehicle Routing Problem (MDVRP).
The node set $V$ in graph $G$ is further partitioned into two subsets: depot nodes $V_d=\{v_1,v_2,\ldots,v_n\}$ and goal nodes $V_g=\{v_{n+1},v_{n+2},\ldots,v_{n+m}\}$. Each edge that belongs to the edge set $E$ has associated distance $c_{ij}$. Let $x_{ijk}$ equal 1 if edge $(i,j)$ is visited by vehicle $k$, and 0 otherwise. We denote an auxiliary variable $y_{ik}$ to set up the sub-tour elimination constraints\\

The numerical model is shown as follows:
\begin{equation}
% {\operatorname{minimize}} \ \tau
{\operatorname{minimize}} \ \sum_{k \in V_d}\sum_{i \in V} \sum_{j \in V} c_{i j} x_{i j k}
\end{equation}
\small
Subject to
\begin{gather}
% \sum_{i \in V} \sum_{j \in V} c_{i j} x_{i j k} \leq \tau \quad \forall k \in V_d\\
\sum_{i \in V} \sum_{k \in V_d} x_{i j k}=1 \quad \forall j \in V_g \label{forvg}\\
\sum_{j\in V} \sum_{k \in V_d} x_{ijk}=1 \quad \forall i \in V_d \label{forvd}\\
\sum_{i\in V} x_{ijk}-\sum_{i \in V} x_{jik}=0 \quad \forall
k \in V_d, j \in V_g \label{flowconserv}\\
\tiny
y_{ik}-y_{jk}+\Tilde{M} x_{i j k} \leqslant \Tilde{M}-1 \quad  \forall (i,j) \in E,  k \in V_d \label{subtour}\\
\small
x_{ijk} \in\{0,1\} \quad \forall i \in V,j \in V, k \in V_d
\end{gather}
\normalsize
% where constraints (2) ensure that $\tau$ equals the maximum distance of all the routes, and 

Constraints (\ref{forvg}) and (\ref{forvd}) ensure that an robot visits each goal location exactly once. Constraints (\ref{flowconserv}) and (\ref{subtour}) are the flow conservation and sub-tour elimination constraints, respectively. \\
In this work, the cost matrix of $c_{ij}$ is created by calculating the distance between vertex $v_i$ and vertex $v_j$, $ i,j \in V$ with $A^*$ planner. Since robots are not required to return to the original position, we simply set the distance from all vertex to depot vertex as 0, which turns depots of robots to dummy destinations that will have no effect on optimal routes.\\
The MD-VRP problem is known to be NP-hard in combinatorial optimization, so to find an assignment and path solution that is close to optimal, the simulated annealing algorithm (\citeauthor{kirkpatrick1983optimization}, \citeyear{kirkpatrick1983optimization}) is employed as a meta-heuristic method to find a near-optimized assignment.\\
An initial solution is obtained using Parallel Greedy Insertion heuristic (\cite{laporte2002classical}), which inserts the cheapest node at its cheapest position iteratively until all nodes are inserted. 
% Inspired by annealing in metallurgy, where a material is heated and then lowered slowly cooled to a state of minimal energy, simulated annealing gradually decreases the probability of accepting worse solutions as the solution space is explored, so as to escape some local minimums. Here, the simulated annealing algorithm starts with the initial route solution described earlier, and in the iteration process, it does not only accepts better solution, but also worse solution decided by a so called Metropolis criterion:\\
% \small
% \begin{equation}
% p=\left\{\begin{array}{cl}
% 1 & \text {, if } f\left(x_{\text {new }}\right) \leq f\left(x\right) \\
% \exp \left(-\frac{f\left(x_{\text {new }}\right)-f\left(x\right)}{T}\right) & \text {, if } E\left(x_{\text {new }}\right)>f\left(x\right)
% \end{array}\right.
% \end{equation}
% \normalsize
% \\
% where $f$ is the maximum route length of all robots. $T$ is "temperature", which decays at a  speed of $T=q \times T$ from an initial temperature $T_i$. $q$ is the decay rate that determines the reduction in $T$ in iterations. The algorithm will terminate in a reasonable length of time or a specified number of solutions are found.
By gradually decreasing the acceptance probability of worse solutions, simulated annealing allows the algorithm to escape local minimums and explore a larger solution space. The threshold acceptance method, proposed by \cite{dueck1990threshold}, is a commonly used cooling schedule in simulated annealing (\citeauthor{santini2018comparison}, \citeyear{santini2018comparison}). It sets a threshold $T$ for accepting worse solutions, which decreases over time until it reaches 0. The algorithm always accepts better solutions but also accepts worse solutions if their gap with the best-found solution is within the threshold $T$.\\
In the solution searching phase, we apply local search heuristics to improve the solutions. The \textit{relocate} method tries to re-insert a node into a new position, while the \textit{SWAP} method switches the positions of two tasks. These methods help to refine the solution obtained from the initial solution found by the Parallel Greedy Insertion heuristic.\\
The SA stops when a predefined maximal iteration, $MAXITER$, is reached. The choice of $MAXITER$ is a trade-off between computational time and solution quality. A larger $MAXITER$ may lead to better solutions but requires more computational resources.\\

Algorithm ~\ref{SA} shows the pseudo-code of the SA. 

\begin{algorithm}[!htbp]

\caption{Simulated Annealing}
 \hspace*{\algorithmicindent} \textbf{Input: } \text{$s$: The initial solution}\\ 
\hspace*{\algorithmicindent} \textbf{Output: } \text{$s^*$: The best found solution}
\begin{algorithmic}[1]
\State{$s^* = s$;}
\REPEAT
\State{$s' = s$;}
\State{use local search to update $s'$;}
\IF{$\displaystyle\frac{f(s')-f(s^*)}{f(s^*)} < T $}
\State{$s = s'$;}
\IF{$f(s') < f(s^*)$}
\State{$s^*=s'$;}
\ENDIF
\ENDIF
\State{$T = T - (T_{initial}/MAXITER)$;}
\UNTIL{$MAXITER$ is met}
\end{algorithmic}
\normalsize
\label{SA}
\end{algorithm}

\subsection{Recurrent Conflict-Based Search (reCBS)}
CBS (\citeauthor{sharon2015conflict}, \citeyear{sharon2015conflict}) is an optimal and complete two-level algorithm that finds non-conflicting paths for each robot starting from their start locations and ending up in goal locations. At the high level, CBS searches the constraint tree, each node of which contains a set of constraints for each robot, the path of each robot found by the low-level search that is consistent with its constraints, and the sum of costs over all the robots. Given the constraints of a node $N$, a low-level search is invoked to return the constraints-consistent time-optimal paths, followed by a new round of conflict detection. To resolve a conflict and guarantee optimality, CBS generates two successor nodes, one of which restricts one robot and unfetters the other robots, the other node contrariwise.
At the low-level, a time-space $A^*$ search is conducted for each robot that satisfies its constraints generated at the high level while completely ignoring other robots. \\
To overcome the limitation of CBS where each robot can only have one start and goal vertex, a recurrent CBS strategy is proposed. In this approach, each robot is assigned a temporary start location ($S_{temp}$) and a temporary goal location ($G_{temp}$) in the initial CBS round. The output of the first round of CBS consists of the conflict-free paths of every robot, and the robot with the shortest path is the first to reach its temporary goal location. All the paths are sliced to match the length of the shortest path, and CBS is called again, taking the current location of each robot as the new start location. For the robot that reaches its temporary goal location the fastest, its temporary goal is updated to its next goal to be visited, while the goal locations for the other robots remain the same. This process is repeated until all the  goal locations are visited, and the sliced paths are appended sequentially to provide a conflict-free path for the robot to visit all assigned goal locations. The robots that have reached their last assigned goal locations are labeled as ``done'' and are included in the next round of CBS in case they block other robots. However, they will not be considered the fastest to reach the temporary goal to avoid the algorithm falling into a dead loop. Additionally, there may be cases where there are redundant robots that are not assigned any tasks. Such scenarios are also taken into account during the implementation of the algorithm. Details are illustrated in algorithm~\ref{recurrent}.

% \begin{algorithm}
% \small
% \caption{Reccurent CBS}
%  \hspace*{\algorithmicindent} \textbf{Input : } SA asignment result\\ 
% % \hspace*{\algorithmicindent}\hspace*{1.3cm}$T$ : List that store all goal positions\\
%  \hspace*{\algorithmicindent} \textbf{Output : } conflict-free path for each robot
% \begin{algorithmic}[1]
% \STATE Declare $\beta$, cost matrix of $N$ by $N$ doubles 
% \IF{$M>N$}
% \STATE call Algorithm~\ref{alpha}
% \STATE call Algorithm~\ref{hierarchical clustering}
% \FORALL{$ i=1$ \textbf{to} $N$}
% \FORALL{$ j=1$ \textbf{to} $N$}
%  $\beta(i,j)= EuclideanDistance(A(i),C(j)$)
% \ENDFOR
% \ENDFOR
% \ELSIF{$M=N$}
% \FORALL{$ i=1$ \textbf{to} $N$}
% \FORALL{$ j=1$ \textbf{to} $N$}
%  $\beta(i,j)= D^*_+.ComputeCost(A(i),T(j)$)
% \ENDFOR
% \ENDFOR
% \ENDIF
% \end{algorithmic}
% \normalsize
% \label{beta}
% \end{algorithm}

\begin{algorithm}[!htbp]
\caption{Reccurent CBS}
\begin{algorithmic}[1]
\STATE {\textbf{Input: } SA assignment result}
\STATE Start locations of each robot
\STATE \textbf{Output: } $P:$ conflict-free paths for each robot
\STATE{$S_{temp} \leftarrow$ start locations of each robot }
\STATE{Initialize the index of the task to be done in assigned tasks as 1 for all robots ($Idx_i=1$ )}
\STATE{Initialize empty list $P$}
\WHILE{True}
\FORALL{robot}
\IF{robot $i$ is not assigned any task}
\STATE{Label $i$ as \textit{done}}
\STATE{$G_{temp,i} \leftarrow$ location of robot $i$}
\ELSE
\STATE{Label $i$ as \textit{working}}
\STATE{$G_{temp,i} \leftarrow$ location of the $Idx_i$ assigned task of robot $i$}
\ENDIF
\ENDFOR
\STATE{$P_{temp}=$ \textbf{CBS} $(S_{temp},G_{temp})$ }
\IF{all robots are labeled as \textit{done}}
\STATE{Append $P_{temp}$ to $P$}
\STATE{ \textbf{break} }
\ELSE
\STATE{In $P_{temp}$, find the \textit{working} robot $i$ that has the shortest length $t$ of path }
\STATE{Slice $P_{temp}$ for all \textit{working} robots to length $t$}
\STATE{Extend $P_{temp}$ for all \textit{done} robots to length $t$ with their last locations}
\STATE{$S_{temp} \leftarrow$ last value of $P_{temp}$}
\STATE{Append $P_{temp}$ to $P$}
\IF{$Idx_i <$  No. assigned tasks of robot $i$}
\STATE{$Idx_i=Idx_i+1$}
\ELSE
\STATE{Label $i$ as \textit{done}}
\ENDIF
\ENDIF
\ENDWHILE
\STATE{\textbf{return} $P$}

\end{algorithmic}
\normalsize
\label{recurrent}
\end{algorithm}
We now discuss the optimality of the proposed algorithm. \\
The SA-reCBS is complete: the simulated annealing will always return a feasible solution; iterations in the recurrent CBS do not affect the completeness of CBS, which has been proved in (\citeauthor{sharon2015conflict}, \citeyear{sharon2015conflict}). Thus, the cascaded SA and recurrent CBS promises to find a solution if one exists. \\
Obviously, the SA-reCBS cannot guarantee an optimal solution for the MRTAPF problem because the simulated annealing in the first layer is a meta-heuristic searching algorithm.

\section{Experiments}
\begin{table*}[t]
\centering
\begin{tabular}{ |p{1cm}||p{2cm}|p{2cm}|p{1cm}||p{2cm}|p{2cm}|  }
  \hline
 % \multicolumn{6}{|c|}{Total cost} \\
 % \hline
Index & SA-reCBS & CBSS & Index & SA-reCBS & CBSS\\
 \hline
1 & 235 & Inf & 11 & 207 & 214\\
 \hline
2 & 187 & 244 & 12 & 215 & 242 \\
 \hline
3 & 190 & 208 & 13 & 197 & 221 \\
 \hline
4 & 184 & 227 & 14 & 211 & 220 \\
 \hline
5 & 224 & 279 & 15 & 215 & 249 \\
 \hline
6 & 249 & 266 & 16 & 202 & 231 \\
 \hline
7 & 196 & 215 & 17 & 211 & 267 \\
 \hline
8 & 217 & 243 & 18 & 207 & 213 \\
 \hline
9 & 215 & 271 & 19 & 229 & 240 \\
 \hline
10 & 215 & Inf & 20 & 201 & 239 \\
 \hline
\end{tabular}
\caption{Comparison of the total travel cost of CBSS and SA-reCBS in 20 instances of map with $n=5$ $m=30$}
\label{tab:costdiff}
\end{table*}
The proposed scheme was implemented on a 1.9GHz AMD Ryzen 7 Pro 5850U laptop with 16GB RAM and evaluated on a total of 480 instances, consisting of 40 instances for each combination of the number of robots $n \in \{ 5,10,20 \}$ and the number of goal locations $m \in \{10,20,30,40\}$, on a 32x32 grid map with 40\% obstacles. The obstacles occupied approximately 40\% of the grids, sometimes slightly less due to overlap. The initial locations of robots, tasks, and obstacles were randomly generated for each instance. The run time of the simulated annealing stage and conflict-based search stage were evaluated separately, along with the total travel distance of each robot.\\
For comparison, we also adapted the CBSS algorithm proposed by \citeauthor{ren2022conflict} (\citeyear{ren2022conflict}) to solve the MRTAPF problem and ran both SA-reCBS and CBSS on the same map under the same conditions (i.e., same number and locations of robots and tasks).
% \citeauthor{ren2022conflict} (\citeyear{ren2022conflict}) proposed the CBSS (conflict-based steiner search) algorithm for the MSPF (multi-goal sequencing path-finding) problems.  Similar to the MRTAPF problem,  MSPF requires every target to be visited at least once and robots to terminate at a unique destination additionally. Here, we adapt the CBSS algorithm to solve MRTAPF problem as the baseline for comparison. We run SA-reCBS and CBSS on the same map with the same conditions (numbers and locations) of robots and tasks and compare the results and computational times.

\begin{figure}[H]
\centering
    \includegraphics[width = 1\linewidth]{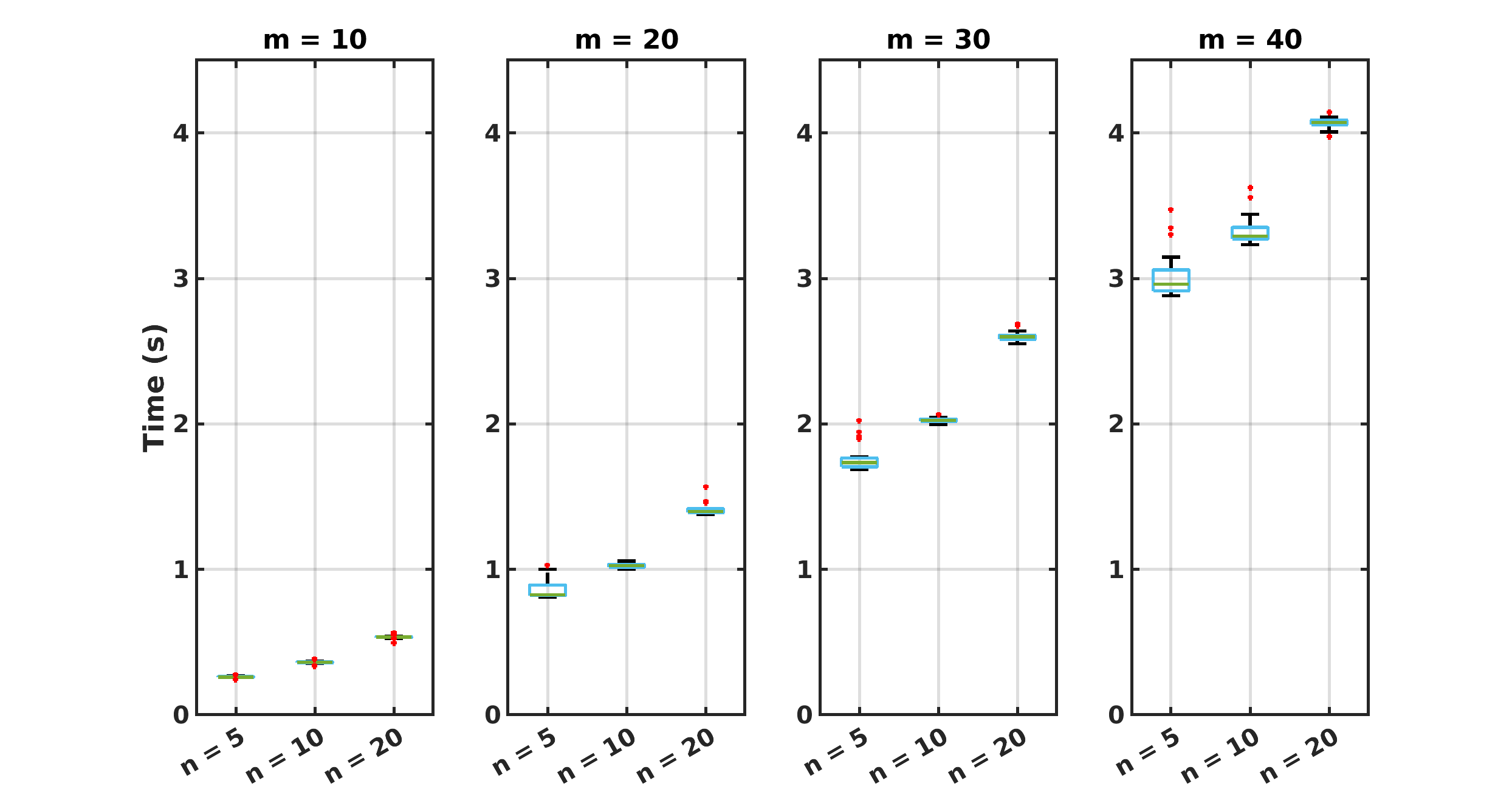}
\caption{Run time statistics for the Simulated Annealing algorithm}
\label{fig:SA}
\end{figure}

\begin{figure}[H]
\centering
    \includegraphics[width = 1\linewidth]{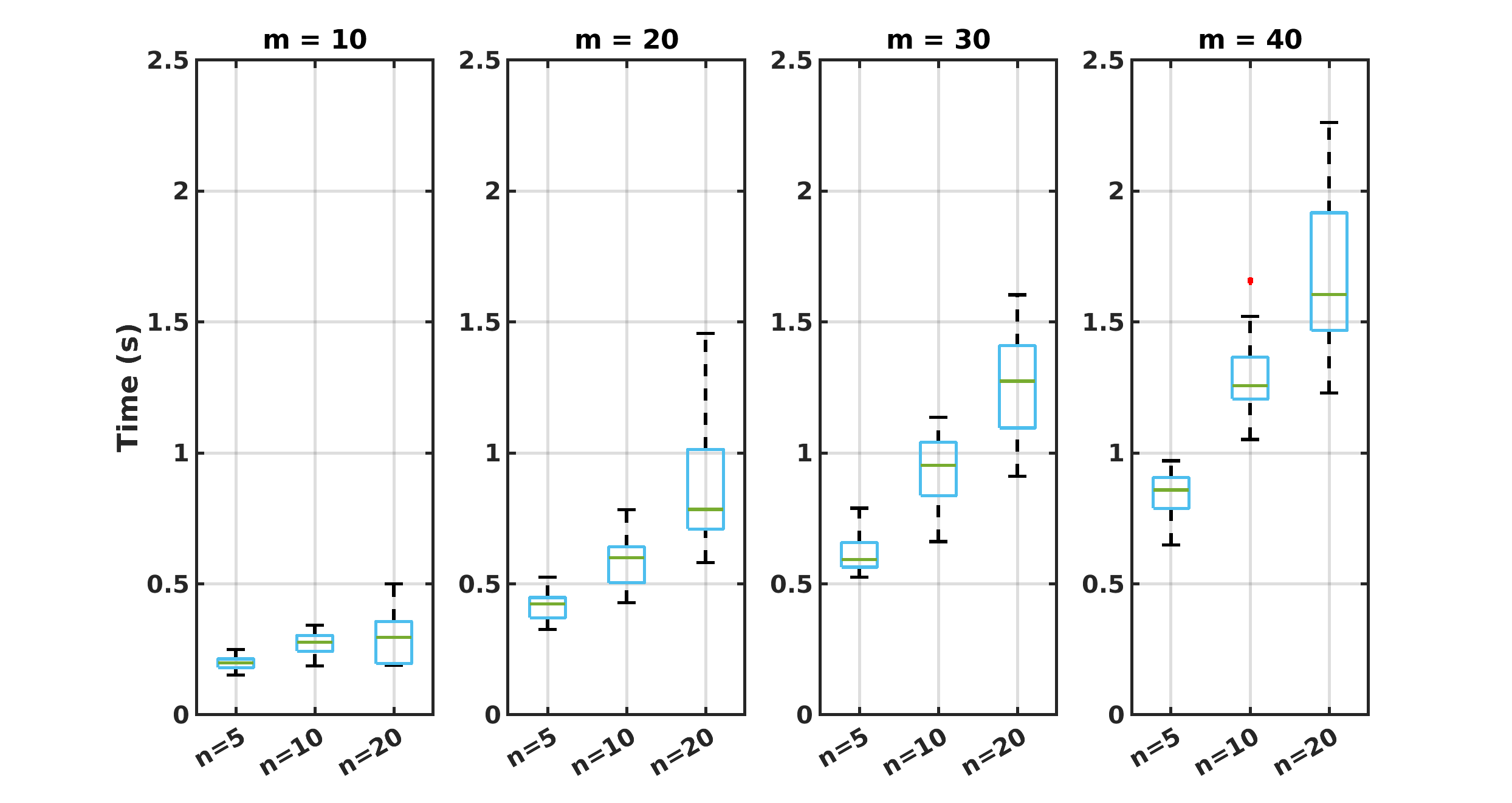}
\caption{Run time statistics for the Recurrent CBS solver}
\label{fig:reCBS}
\end{figure}

\begin{figure}[H]
\centering
    \includegraphics[width = 1\linewidth]{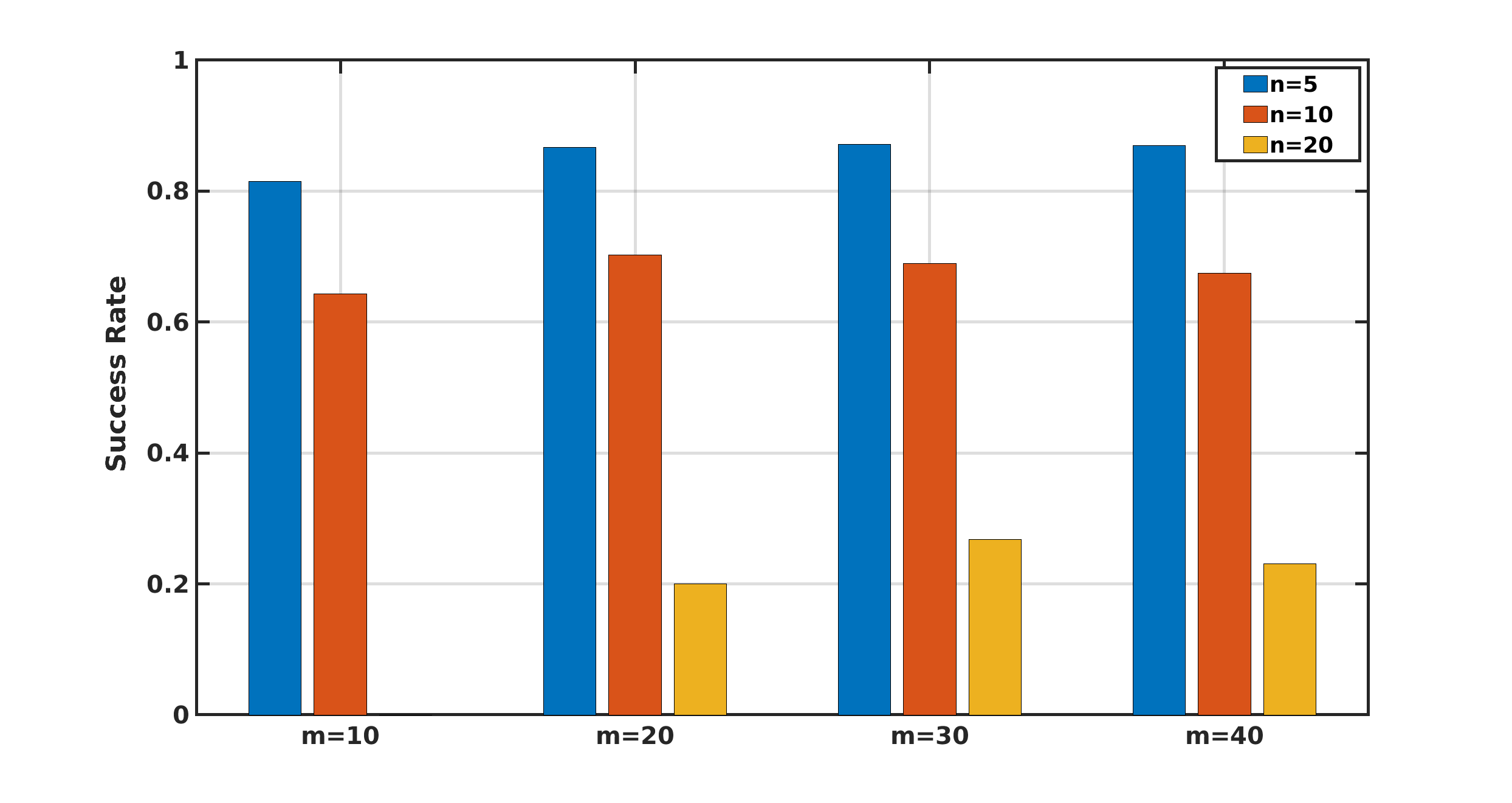}
\caption{Success rate of CBSS within 60 seconds}
\label{fig:success rate}
\end{figure}

\begin{figure}[H]
\centering
    \includegraphics[width = 1\linewidth]{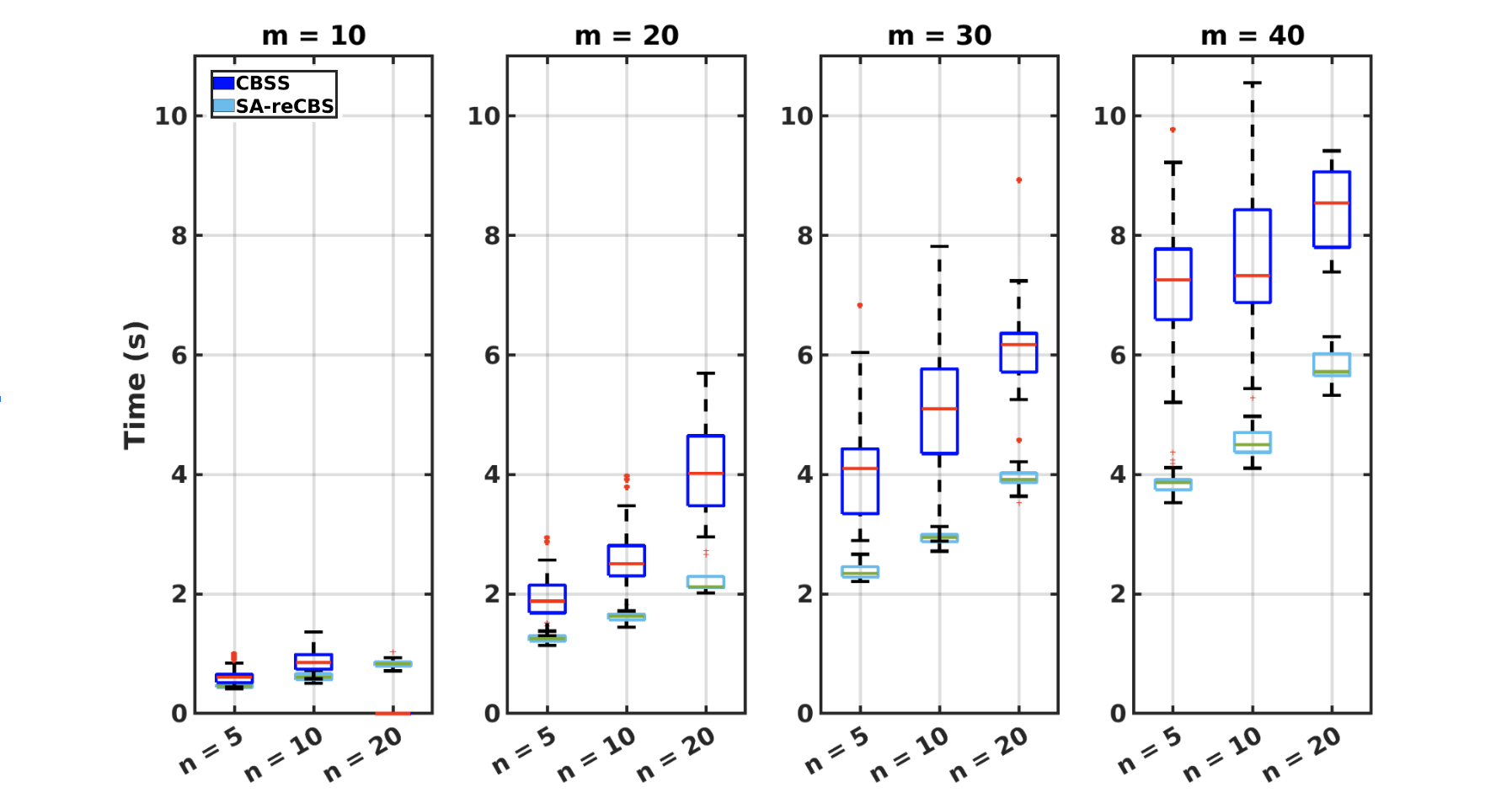}
\caption{Execution time of CBSS (success) and SA-reCBS}
\label{fig:timediff}
\end{figure}

\begin{figure}[H]
\centering
    \includegraphics[width = 1\linewidth]{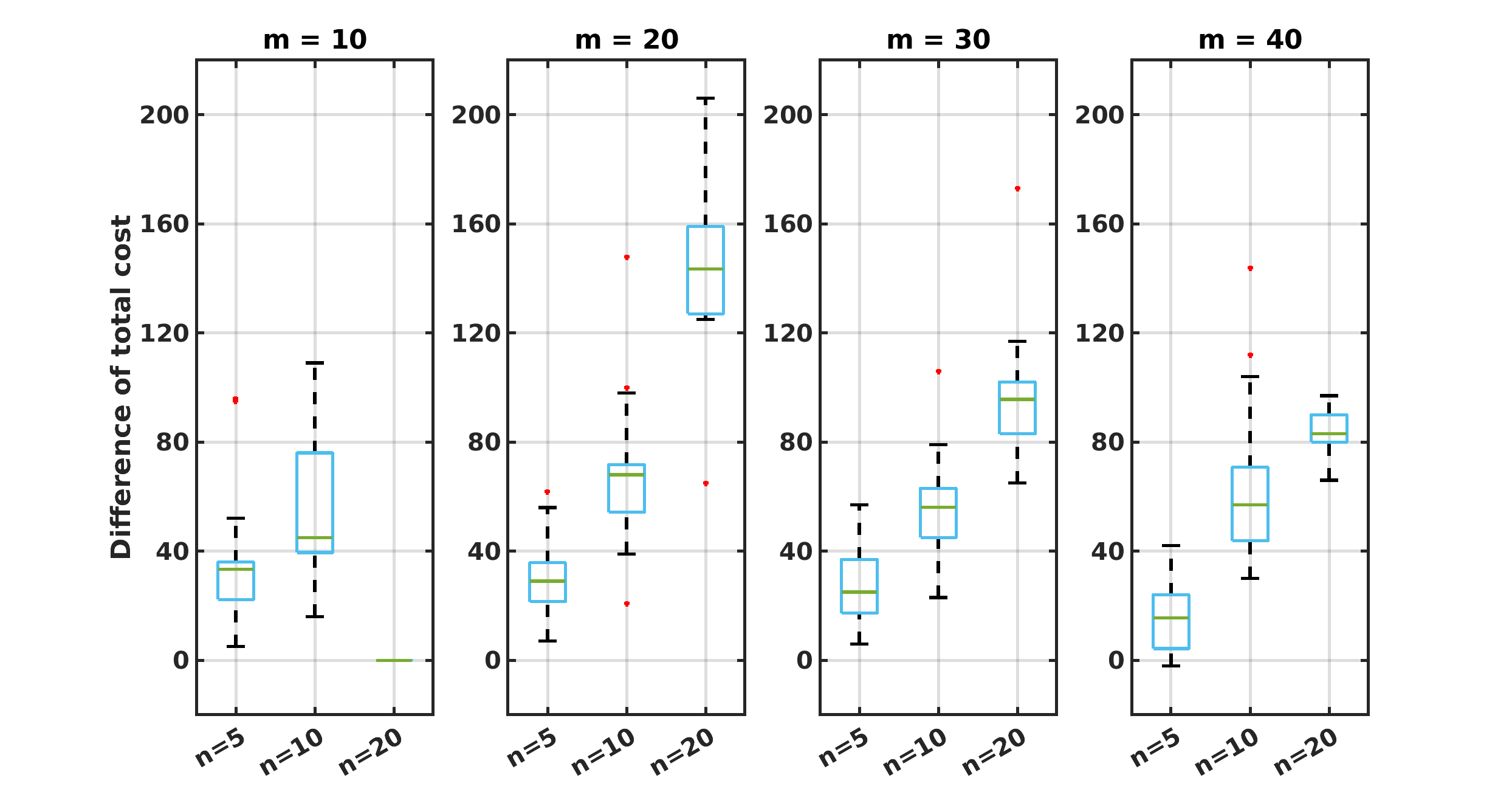}
\caption{The difference of total cost of CBSS and SA-reCBS}
\label{fig:costdiff}
\end{figure}

SA-reCBS consists of two stages, the simulated annealing for routing and the recurrent conflict-based search for collision avoidance. The computational time for the two stages was calculated separately and is presented in Figure~\ref{fig:SA} and Figure~\ref{fig:reCBS}, respectively. The statistical characteristics of the computational results were shown using boxplots, which illustrate the range of values, as well as the lower and upper quartiles and median values. In the boxplots, the line at the bottom and top represents the minimum and maximum values, respectively, while the bottom and top of the box indicate the Q1 and Q3 points, respectively. Any red points outside of the box-whisker section represent outlier values.\\ 
Figure~\ref{fig:SA} depicts the run time statistics of the assignment-routing stage. As the number of robots and tasks increases, the complexity of the problem rises. Therefore, the computational time of the simulated annealing algorithm increases accordingly. \\
The execution time of the recurrent CBS, displayed in Figure~\ref{fig:reCBS}, rises gradually as the number of robots and tasks grows, from around 0.2$s$ with 5 robots and 10 tasks to 1.6$s$ with 20 robots and 40 tasks. This is expected as the iteration times of CBS increase with the advance of the number of robots and tasks.\\
Figure~\ref{fig:success rate} shows the success rate of CBSS for the instances. The success rate refers to the rate of finding feasible solutions within a time limit of 60 seconds. The data for $n=20$, $m=10$ is missing because CBSS requires the number of tasks greater than the number of robots. From the bar chart, we can say that the CBSS algorithm performs similarly for the same number of robots, regardless of the number of tasks, but drops drastically in performance as the number of robotsgoes up (20\% success rate with 20 robots). Whereas SA-reCBS remains a 100\% success rate (not shown in the figure). \\
To make a fair comparison between CBSS and SA-reCBS, the instances where CBSS is able to find a solution are sifted out, and the time efficiency and total cost of the two algorithms are compared.  The computational time of both CBSS and SA-reCBS (the sum of the run time of SA and recurrent CBS) increases monotonically as the total number of robots and tasks increases, as shown in Figure~\ref{fig:timediff}. Apart from the instance of $n=20$, $m=10$, where CBSS is not able to find a solution, SA-reCBS consistently outperforms CBSS in terms of computational time. In Figure~\ref{fig:costdiff}, we subtracted the total cost of SA-reCBS from that of CBSS to compare the optimization result of the two algorithms. In most cases, SA-reCBS found paths for robots with a shorter total distance, as indicated by the fact that nearly all the subtraction values were positive. However, we are aware that CBSS found better results in rare scenarios, such as when $n=5$ and $m=40$, as the minimum value of the subtraction was below 0 in the fourth sub-figure. Additionally, table~\ref{tab:costdiff} presents the total travel cost of all robots of CBSS and SA-reCBS as an example.\\
Overall, for MRTAPF problem, the proposed SA-reCBS outperforms CBSS in terms of computational time, optimization results and success rate, especially the success rate. On the one hand, the transformation method for solving mTSP in CBSS (\citeauthor{ren2022conflict}, \citeyear{ren2022conflict}) is less effective in solving the MDVRP problem. On the other hand, CBSS couples CBS and mTSP to find a local optimal solution with a conflict search forest, resulting in a long computational time.\\

\section{Conclusion}
In this work, we formulated a novel two-layer architecture for multi-robot task allocation and reactive path-finding that aims to minimize the total distance traveled by the robots. The proposed SA-reCBS integrates the simulated annealing algorithm with a recurrent conflict-based search to assign tasks and plan collision-free paths for multiple robots. SA-reCBS has a notable advantage over other solvers for vehicle-routing problems and multi-agent path-finding (MAPF) problems since it assigns all tasks to robots without grouping them based on the number of robots, while also guaranteeing no robot collisions. The performance of the architecture is evaluated across several grid map instances to assess its computational efficiency. In comparison to cutting-edge algorithm CBSS, SA-reCBS typically yields solutions with lower computational time and total costs.\\
%\begin{ack}
%Place acknowledgments here.
%\end{ack}

\bibliography{ifacconf}             % bib file to produce the bibliography
\end{document}